\documentclass{article}
\usepackage{ijcai24}
\usepackage{color}
\usepackage{times}
\usepackage{soul}
\usepackage{url}
\usepackage[hidelinks]{hyperref}
\usepackage[utf8]{inputenc}
\usepackage[small]{caption}
\usepackage{graphicx}
\usepackage{amsmath}
\usepackage{amsthm}
\usepackage{booktabs}
\usepackage{multirow}
\usepackage{algorithm}
\usepackage{algorithmic}
\usepackage[switch]{lineno}
\usepackage{amssymb}

\title{1DFormer: a Transformer Architecture Learning 1D Landmark Representations for  Facial Landmark Tracking}

\author{
	Shi Yin$^{1,2}$
	\and
	Shijie Huang$^1$\and
	Shangfei Wang $^{3}$\and
	Jinshui Hu*$^1$ \and
	Tao Guo$^1$ \and
	Bing Yin$^1$ \and
	Baocai Yin$^1$ \And
	Cong Liu*$^1$ 
	\affiliations
	$^1$iFLYTEK Research\\
	$^2$Institute of Artificial Intelligence, Hefei Comprehensive National Science\\
	$^3$University of Science and Technology of China 
	\emails
	*Correspondence to jshu@iflytek.com and congliu2@iflytek.com.
}

\begin{document}

\maketitle

\begin{abstract}
	Recently, heatmap regression methods based on 1D landmark representations have shown prominent performance on locating facial landmarks.  However, previous  methods ignored to make deep explorations on the good potentials of 1D landmark representations for sequential and structural modeling of multiple landmarks to track facial landmarks. To address this limitation, we propose a Transformer architecture, namely 1DFormer, which learns informative 1D landmark representations by capturing the dynamic and the geometric patterns of landmarks via token communications in both temporal and spatial dimensions for facial landmark tracking. For temporal modeling, we propose a confidence-enhanced multi-head attention mechanism with a recurrently token mixing strategy to adaptively and robustly embed long-term landmark dynamics into their 1D representations; for structure modeling, we design intra-group and inter-group geometric encoding mechanisms to encode the component-level as well as global-level facial structure patterns as a refinement for the 1D representations of landmarks through token communications in the spatial dimension via 1D convolutional layers. Experimental results on the 300VW and the TF databases show that 1DFormer successfully models the long-range sequential patterns as well as the inherent facial structures to learn informative 1D representations of landmark sequences, and achieves state-of-the-art performance on facial landmark tracking.
\end{abstract}

\section{Introduction}
\begin{figure}[t]
	\centering
	\includegraphics[scale=0.3]{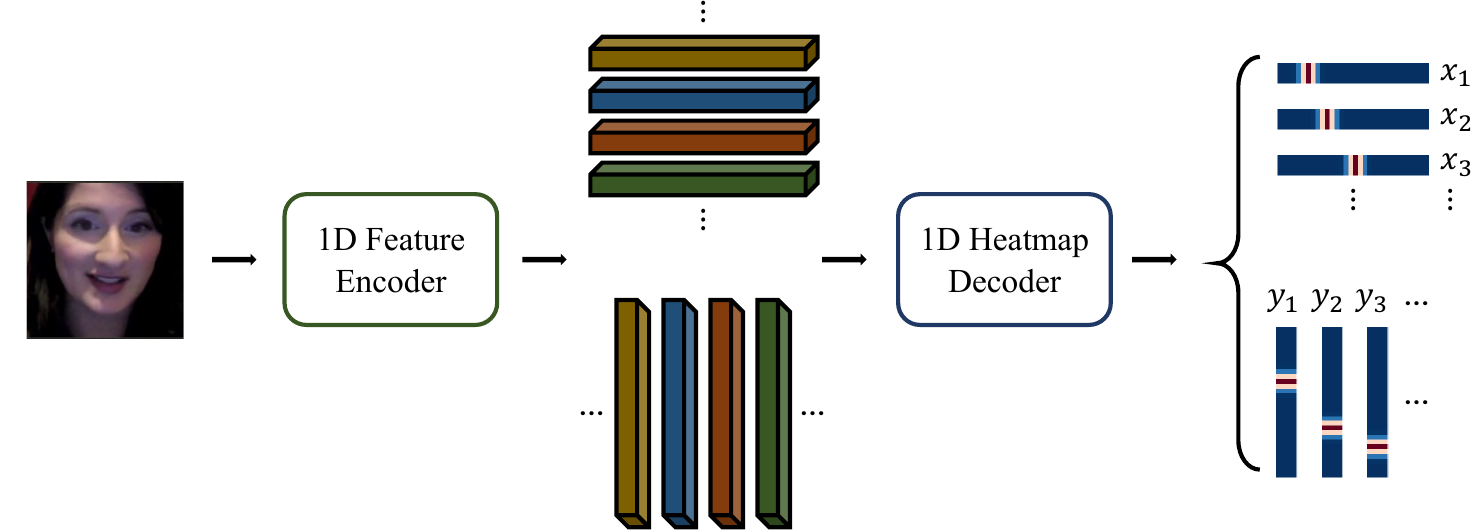}
	\caption{An illustration of 1D heatmap regression methods, which are built upon 1D landmark representations, including 1D feature vectors and heatmaps. Although current methods achieved remarkable performance on detecting landmarks, they ignored a deep exploration on temporal and structural modeling, which is critical for landmark tracking.}
	\label{DRL}
\end{figure}

Tracking facial landmarks from a video stream \cite{survey,survey2} is a fundamental task for human-centered applications, such as emotion analysis and human-computer interaction, while remaining unsolved in the challenging ``in-the-wild" scenarios. Recently, 1D heatmap regression methods, which are built based on 1D landmark representations, i.e., the light-weight 1D feature vectors and heatmaps representing the marginal distribution of every landmark on each axis as depicted in Figure \ref{DRL}, have achieved prominent performance for landmark localization of human faces \cite{AOHR,DBLP:conf/accv/XiongZDS20,HybridMatch}, human bodies \cite{I2L,SimCC,chi2022human} and objects \cite{DBLP:journals/remotesensing/LiuZCW22}.  Compared to the 2D landmark representations  \cite{hourglass,how_far,HRnet} with quadratic spatial complexity, the 1D representations, with a linear spatial complexity, could achieve higher resolution on each axis and enough feature channels under the limited machine memory. Thus, these methods based on 1D representations could capture more details of the spatial patterns of landmarks and bring a lower quantization error as well as a better accuracy on locating facial landmarks under the same hardware condition. However, previous methods have not fully explored the good potentials of 1D landmark representations for temporal sequence modeling, which is critical for the facial landmark tracking task. What's more, all of these methods predicted the 1D representation of each landmark in a loosely coupled manner, lacking considerations of structural modeling upon multiple landmarks. These limitations caused the accuracy bottleneck of these methods.

\begin{figure*}
	\centering
	\includegraphics[scale=0.37]{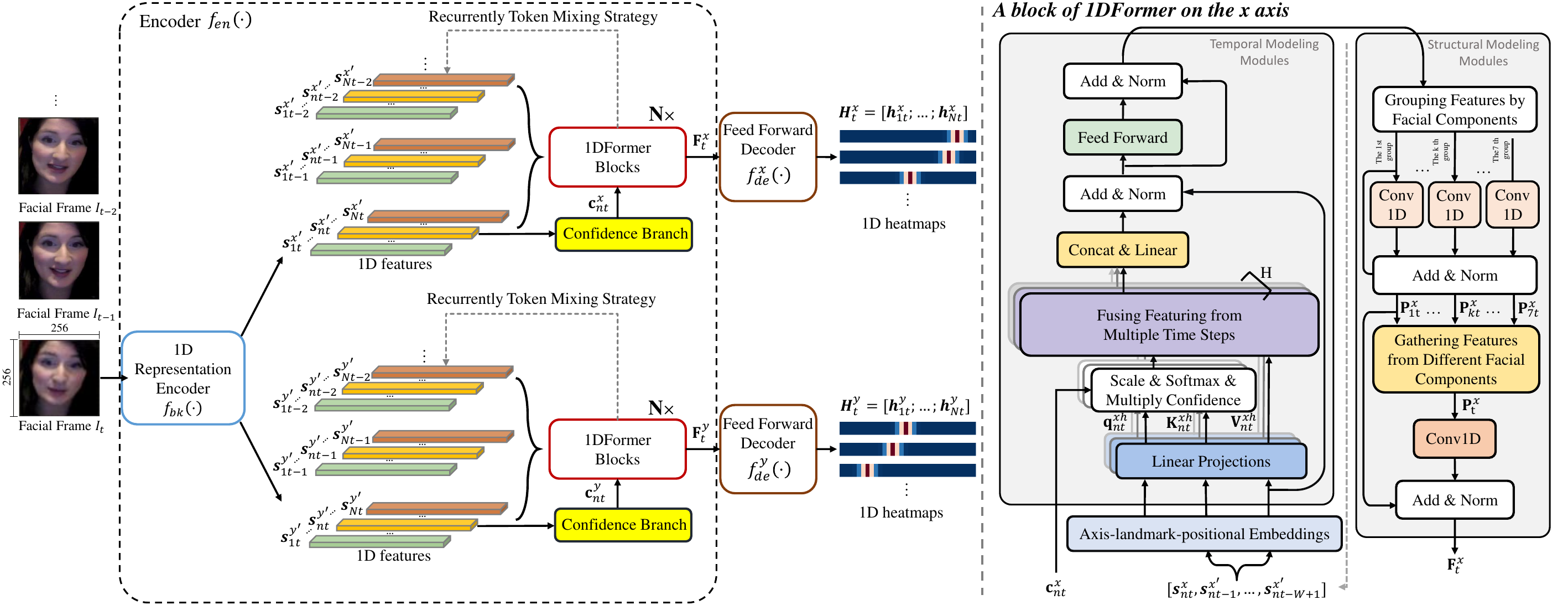}
	\caption{Upper part: an architectural overview of the proposed facial landmark tracking method, i.e., 1DFormer. Lower part
		: the internal architecture of a basic block of 1DFormer on the x axis. The architecture of basic block on the y axis is the same as the x axis.
	}
	\label{framework}
\end{figure*}

To address these, as shown in Figure \ref{framework}, we propose a Transformer-based 1D representation learning method, i.e., 1DFormer, to effectively capture the temporal and structural patterns of multiple landmarks for facial landmark tracking.  The clusters of 1D landmark representations, including the hidden features as well as the output heatmaps, are in the form of a sequence of 1D light-weight vectors, which are the ideal input and output patterns of Transformer. Thus, we build our method based on the paradigm of Transformer and develop a Transformer architecture with both temporal modeling and structural modeling techniques. For temporal modeling, we propose a confidence-enhanced multi-head attention mechanism with a recurrently token mixing strategy to integrate long-range temporal information from past frames to enhance the features of the current time step through efficient window sliding and information delivery. The attention mechanism combines the temporal correlations learned from the key-query mechanism as well as the  confidence score of feature qualities as the integrated attention weights, robustly guiding the fusion of features from different time steps despite challenging imaging conditions. For structural modeling, to make full use of the intrinsic geometry patterns of faces for landmark tracking, we design an intra-group and an inter-group geometry encoding mechanisms to embed component-level as well as global-level facial structure patterns as a refinement for the 1D representations of the landmark sequence. It is expected that these mechanisms can effectively extract the structural patterns of multiple landmarks, and meanwhile, keep the clear semantic of each landmark. For that purpose, we adopt 1D convolutional layers instead of the vanilla attention layers for token communications in the spatial dimension. Through the structural modeling mechanisms, the learned facial structures are helpful to correct the texture disturbances caused by occlusions or uneven illumination conditions, and further improve the landmark tracking performance for ``in-the-wild" scenarios.

The main contributions of this paper can be summarized as the following. First, as far as we know, we are the first to fit the prominent Transformer paradigm on learning informative 1D representations of facial landmark sequences for facial landmark tracking. Second, we propose a new Transformer architecture, namely 1DFormer, with both temporal modeling and structural modeling mechanisms, to explore the good potentials of 1D representation learning on modeling the long-term sequential as well as the geometric patterns of facial landmarks. Third, as demonstrated by experiments on the 300VW database and the TF database, the tracker based on 1DFormer achieves state-of-the-art performance for facial landmark tracking with a significant improvement of accuracy and stability performance compared to the related works.

\section{Related Work}
\label{related_work}

As the mainstream of key point detection and tracking, heatmap regression methods explicitly modeled the spatial distributions of landmarks by learning to predict the probability intensity that a landmark appears in each position of the 2D plane or 3D space. Through explicitly capturing the landmark distributions, these methods achieved a better spatial generalization performance on facial landmark detection and tracking compared to the conventional coordinate regression methods which directly regressed the values of landmark coordinates \cite{SDM,TSTN,TSCN,GAN_Tracking,DBN,RBM,Qtracker}. Heatmap regression methods can be further divided into two categories: 2D heatmap regression and 1D heatmap regression. 

2D heatmap regression methods \cite{hourglass,encoder_decoder,how_far,HRnet,CascadedTransformers,RePFormer} modeled 2D joint distributions, typically assumed as 2D Gaussian distributions,  on the x and y axes in both hidden and output stages for each landmark. These approaches excelled in spatial generalization but faced challenges due to the high spatial complexity, i.e., $\mathcal{O}(L^2)$ where $L$ is the resolution of 2D features and heatmaps, requiring down-sampling of feature and heatmap resolutions under limited machine space. This leaded to precision loss and quantization errors, thus limiting the effectiveness in landmark detection  tracking. To address this, 1D heatmap regression methods \cite{AOHR,DBLP:conf/accv/XiongZDS20,I2L,SimCC,chi2022human,DBLP:journals/remotesensing/LiuZCW22} replaced some 2D landmark representations of high spatial complexity with 1D features and heatmaps. These methods, assuming 1D Gaussian distribution as the  marginal distribution of landmarks on each axis, used strided CNN \cite{AOHR}, MLP \cite{SimCC}, or pooling \cite{I2L} to compress image representations along one axis while enhancing resolution along the other. With a feature resolution of $L$, the spatial complexity is reduced to $\mathcal{O}(L)$, allowing higher resolutions per axis and more feature channels to capture fine-grained details of landmark distributions. This brings improved landmark localization performance compared to both coordinate and 2D heatmap regression methods.

However, previous 1D heatmap regression methods, mostly detection-based \cite{DBLP:conf/accv/XiongZDS20,I2L,SimCC,chi2022human,DBLP:journals/remotesensing/LiuZCW22,HybridMatch},  lacked depth in time series modeling crucial for landmark tracking.  To the best of our knowledge, only  \citeauthor{AOHR} \shortcite{AOHR} applied 1D representations to model temporal sequences of landmarks. However, their sequential encoder was quite simple with fixed preset weights for fusing features from the past frames, without a dynamic weight assignment mechanism to adaptively model the temporal correlations and confidences of feature qualities from different frames.  These preset weights also decreased quickly with the exponential decay of time, causing a sub-optimal performance on capturing long-term temporal dependencies from historical frames. What's more, these methods ignored to develop an effective interaction mechanism to capture the structural patterns of multiple landmarks. To address these issues, we propose a new Transformer architecture adept at long-range temporal modeling and geometric modeling to enhance 1D representation learning for facial landmark tracking.

\section{Method}
The proposed facial landmark tracker consists of an 1D feature encoder $f_{en}(\cdot)$ as well as two 1D heatmap decoders $f^x_{de}(\cdot)$ and $f^y_{de}(\cdot)$. The 
video to be extracted facial landmarks is denoted as $\textbf{I}_{1:T}=[\textbf{I}_{1};\textbf{I}_{2},...;\textbf{I}_{T}]$, where $\textbf{I}_{t}\in\mathbb{R}^{ H\times W\times 3} (1\leq t \leq T)$ is the image of the $t$ th frame. $f_{en}(\cdot)$ encodes $\textbf{I}_{1:T}$ as a cluster of features representing the marginal distributions of landmarks on the $x$ and $y$ axes for each time step:
\begin{equation}
	\label{en1}
	\textbf{F}^{x}_{1},\textbf{F}^{y}_{1},\textbf{F}^{x}_{2},\textbf{F}^{y}_{2},...,\textbf{F}^{x}_{T},\textbf{F}^{y}_{T} = f_{en}(\textbf{I}_{1:T};\theta_{en})
\end{equation}
where $\theta_{en}$ is the parameters of $f_{en}(\cdot)$, $\textbf{F}^{x}_{t}=[ \textbf{f}^x_{1t};$ $\textbf{f}^x_{2t}; ...; \textbf{f}^x_{Nt}] \in\mathbb{R}^{N \times L}$ and $\textbf{F}^{y}_{t}=[ \textbf{f}^y_{1t};$ $\textbf{f}^y_{2t}; ...; \textbf{f}^y_{Nt}]\in\mathbb{R}^{N \times L}$ respectively denote the 1D representations on $x$ axis and the $y$ axis at the $t$ th time step. Here $N$ is the number of the pre-defined landmarks, $\textbf{f}^x_{nt} (1\leq n \leq N)\in\mathbb{R}^{1 \times L}$ and $\textbf{f}^y_{nt}\in\mathbb{R}^{1 \times L}$ denote the 1D features for the $n$ th landmark at the $t$ th time step. Based on $\textbf{F}^{x}_{t}$ and $\textbf{F}^{y}_{t}$, the fully connected feed forward 1D heatmap decoders $f^x_{de}(\cdot)$ and $f^y_{de}(\cdot)$, respectively predict two groups of 1D heatmaps, i.e., $\textbf{h}^x_{nt}=f^x_{de}(\textbf{f}^x_{nt})\in\mathbb{R}^{1 \times D}$ and $\textbf{h}^y_{nt}=f^y_{de}(\textbf{f}^y_{nt})\in\mathbb{R}^{1 \times D}$, for the two axes, respectively.  The $x$ and $y$ coordinates of the landmark can be obtained from the peak positions of $\textbf{h}^x_{nt}$ and $\textbf{h}^y_{nt}$, respectively. In this paper, we propose a new design of $f_{en}(\cdot)$, which is composed of a backbone encoder extracting 1D representations from each individual frame image, as well as a Transformer, which we called as 1DFormer, to extract long-range sequential patterns and geometric structures of facial landmarks to refine their 1D representations.

\subsection{Backbone Encoder}
\label{be}
Before processed by 1DFormer, we first extract 1D representations for every landmark from each individual frame image through a backbone 1D representation encoder, which is composed of a FAN network \cite{how_far} cascaded with two groups of strided CNNs \cite{AOHR} to respectively encode 1D representations, i.e., $\textbf{s}^x_{nt} (1\leq n \leq N, 1\leq t \leq T)\in\mathbb{R}^{1 \times L}$ and $\textbf{s}^y_{nt}$, on the two axes, respectively.

\subsection{Temporal Modeling Mechanism of 1DFormer}

We take attention mechanism to embed the dynamic patterns of the landmark sequence into $\textbf{s}^x_{nt} (1\leq n \leq N, 1 \leq t \leq T)$ and $\textbf{s}^y_{nt}$, respectively, by integrating the 1D landmark representations from different time steps. The inner structures of our attention modules will be discussed later. Here we firstly present a recurrently token mixing strategy arranging the inputs and outputs of the attention modules to capture the long-range sequential patterns of facial landmarks from the video stream, as shown in Eq. \eqref{transxy}:
\begin{equation}
	\begin{split}
		\label{transxy}
		& \textbf{s}^{x'}_{nt} = Attention(\textbf{s}^{x}_{nt}, \textbf{s}^{x'}_{n(t-1)},...,\textbf{s}^{x'}_{n(t-W+1)})\\
		& \textbf{s}^{y'}_{nt} = Attention(\textbf{s}^{y}_{nt}, \textbf{s}^{y'}_{n(t-1)},...,\textbf{s}^{y'}_{n(t-W+1)})\\
	\end{split}
\end{equation}
where $\textbf{s}^{x'}_{nt}$ and $\textbf{s}^{y'}_{nt} (1\leq n \leq N, 1 \leq t \leq T)$ are the features integrated with temporal information of the video through mixing temporal tokens, i.e., $\textbf{s}^{x}_{nt}, \textbf{s}^{x'}_{n(t-1)},...,\textbf{s}^{x'}_{n(t-W+1)}$ or $\textbf{s}^{y}_{nt}, \textbf{s}^{y'}_{n(t-1)},...,$ $\textbf{s}^{y'}_{n(t-W+1)}$. As a computationally acceptable way, instead of taking the features from all of the past frames as the input, we just send the features from a temporal neighborhood window of $W$ frames into the attention modules to refine the 1D features of landmarks with sequential patterns.  Since the memories of temporal patterns are delivered with the sliding of the window from the past windows to the current window, $\textbf{s}^{x'}_{nt}$ and $\textbf{s}^{y'}_{nt}$ can still keep a long memory of the landmark dynamics.

We extend the conventional positional embedding of Transformer as axis-landmark-positional (alp) embeddings which distinguish not only different position indexes but also different axes and landmarks, to make the attention modules aware of the discrepancy of marginal distributions on different axes, the semantic distinctions among different facial landmarks, as well as the relative position of each feature vector from a window. These embeddings are denoted as $\textbf{e}^{x}_{nw}$ and $\textbf{e}^{y}_{nw} (1 \leq n \leq N, 1 \leq w \leq W)$, respectively, where the superscript ($x$ or $y$) denotes the axis the 1D feature belongs to, subscript $n$ is the landmark index, $w$ is the relative position index of an input feature to the whole input window with $W$ frames.  The input feature sequences for attention modules are added with the axis-landmark-positional embeddings in the way like: $\textbf{s}^{x}_{nt}:=\textbf{s}^{x}_{nt}+\textbf{e}^{x}_{n1}$, $\textbf{s}^{x'}_{n(t-w+1))}:=\textbf{s}^{x'}_{n(t-w+1)}+\textbf{e}^{x}_{nw} (1< w \leq W)$, $\textbf{s}^{y}_{nt}:=\textbf{s}^{y}_{nt}+\textbf{e}^{y}_{n1}$, $\textbf{s}^{y'}_{n(t-w+1))}:=\textbf{s}^{y'}_{n(t-w+1)}+\textbf{e}^{y}_{nw} (1< w \leq W)$. Here $\textbf{e}^{x}_{nw}$ and $\textbf{e}^{y}_{nw}$ are randomly initialized, then optimized with the tracker.

Next, we present the inner structure of our attention mechanism, which specifically is a confidence-enhanced multi-head attention mechanism, to dynamically determine the attention weights of features from different time steps according to their temporal correlations as well as the their qualities influenced  by the imaging conditions. Typically, the attention weights can be calculated by the scaled dot-product operations between the query and the key matrices:
\begin{equation}
	\label{eq:bs1}
	\begin{aligned}
		\textbf{q}^{xh}_{nt} & = \textbf{s}^x_{nt}{\textbf{W}^{xh}_{q}}^T,\,\,\,	\textbf{q}^{yh}_{nt} = \textbf{s}^y_{nt}{\textbf{W}^{yh}_{q}}^T \\
		\textbf{K}^{xh}_{nt} & = \textbf{I}^{x}_{nt}{\textbf{W}^{xh}_{k}}^T,\,\,\,
		\textbf{K}^{yh}_{nt} = \textbf{I}^{y}_{nt}{\textbf{W}^{yh}_{k}}^T \\
		\textbf{a}^{xh}_{nt} & = sf(\frac{\textbf{q}^{xh}_{nt} {\textbf{K}^{xh}_{nt}}^T}{\sqrt{d_h}}),\,\,\,\textbf{a}^{yh}_{nt} = sf(\frac{\textbf{q}^{yh}_{nt} {\textbf{K}^{yh}_{nt}}^T}{\sqrt{d_h}})\\
	\end{aligned}
\end{equation}
where $sf(\cdot)$ denotes the $Softmax$ function; $\textbf{I}^{x}_{nt} = [\textbf{s}^{x}_{nt}; \textbf{s}^{x'}_{n(t-1)};...;$ $\textbf{s}^{x'}_{n(t-W+1)}] \in \mathbb{R}^{W \times L}$ and $\textbf{I}^{y}_{nt} $ denote the feature sequence for the $t$ th input window chunk of the attention module; $h$ denotes the head index of the multi-head attention mechanism; $\textbf{q}^{xh}_{nt}$ and $\textbf{q}^{xh}_{nt}$ are the $h (1\leq h \leq H)$ th query vector pair to get the attention weight vectors, i.e., $\textbf{a}^{xh}_{nt} \in \mathbb{R}^{1 \times W}$ and $\textbf{a}^{yh}_{nt} \in \mathbb{R}^{1 \times W}$, for the $t$ the time step; $\textbf{K}^{xh}_{n}$ and $\textbf{K}^{yh}_{n} \in \mathbb{R}^{W \times d_h}$ are the $h$ th key matrix pair; $\textbf{W}^{xh}_{q}$, $\textbf{W}^{yh}_{q} $, $\textbf{W}^{xh}_{k}$, and $\textbf{W}^{yh}_{k} \in \mathbb{R}^{d_h \times L}$ are the projection matrices. By optimizing the attention mechanism, $\textbf{a}^{xh}_{nt}$ and $\textbf{a}^{yh}_{nt}$ can implicitly capture the correlations among features of the temporal sequence: the intensity of the $w (1 \leq w \leq W)$ th element of $\textbf{a}^{xh}_{nt}$ reflects the correlation between $\textbf{s}^{x}_{nt}$ and $\textbf{s}^{x'}_{n(t-w+1)}$. 

We extend the conventional attention mechanism with two confidence prediction branches $f^x_{c}(\cdot)$ and $f^y_{c}(\cdot)$ to predict the confidence score of feature qualities at each time step. We take $\textbf{s}^x_{nt}$ and $\textbf{s}^y_{nt}$ instead of $\textbf{s}^{x'}_{nt}$ and $\textbf{s}^{y'}_{nt}$ as the input for the confidence predictors to reflect the imaging quality of each individual video frame. It is expected that, if the texture features around the $n$ th landmark at the $t$ th frame are of good quality with a clear positional discrimination for the landmark, $c^x_{nt}=f^x_{c}(\textbf{s}^x_{nt})$ and $c^y_{nt}=f^y_{c}(\textbf{s}^y_{nt})$ are assigned as a high value and the effects of $\textbf{s}^x_{nt}$ and $\textbf{s}^y_{nt}$ on temporal token communications will be encouraged; otherwise, if the image textures of the $n$ th landmark are polluted at the current time step by occlusions, $c^x_{nt}$ and $c^y_{nt}$ are lowered, and $\textbf{s}^x_{nt}$ as well as $\textbf{s}^y_{nt}$ will be mitigated to alleviate the effect of feature disturbances.  We integrate the weights calculated by the key-query operations with the confidence scores, namely $\textbf{c}^x_{nt} = [c^x_{nt}, c^x_{n(t-1)}, \ldots,  c^x_{n(t-W+1)}]$ and $\textbf{c}^y_{nt} \in \mathbb{R}^{1\times W}$, as the final attention weights, to model temporal correlations in landmark sequences while maintaining robustness against texture disturbances, as shown in Eq. \eqref{eq:bs2}:

\begin{equation}
	\label{eq:bs2}
	\begin{aligned}
		\textbf{a}^{xh}_{nt} & = sf(\frac{\textbf{q}^{xh}_{nt} {\textbf{K}^{xh}_{nt}}^T}{\sqrt{d_h}}\odot\textbf{c}^{xT}_{nt}),\,\,\,\textbf{a}^{yh}_{nt} = sf(\frac{\textbf{q}^{yh}_{nt} {\textbf{K}^{yh}_{nt}}^T}{\sqrt{d_h}}\odot\textbf{c}^{yT}_{nt})
	\end{aligned}
\end{equation}

Based on $\textbf{a}^{xh}_{nt}$ and $\textbf{a}^{yh}_{nt}$, the features from different time steps are fused as the temporally refined 1D landmark representations, i.e., 
$\textbf{s}^{x'}_{nt}$ and $\textbf{s}^{y'}_{nt}$, for the $n$ landmark at the $t$ th time step. Since the feature fusion steps are vanilla in Transformer, we just attach its details in the supplementary material due to the page limits.

\subsection{Structural Modeling Mechanism of 1DFormer}
\label{structure}
We embed facial structural patterns to enhance the 1D representation of each landmark, as shown in the right-most part of Figure \ref{framework}. According to the inherent structure of a human face, its landmarks can be divided into seven groups, i.e., left eyebrow, right eyebrow, left eye, right eye, nose, mouth, and contour. 
To encode component-level structural patterns as a refinement for 1D landmark representations, we propose an intra-group geometric encoding mechanism. Specifically, at the $t (1\leq t \leq T)$ th time step, we stack the features outputted by the attention modules for landmarks of the $k  (1 \leq k \leq 7)$ th group as $\textbf{S}^{x'}_{kt}= stack(\textbf{s}^{x'}_{k(1)t},\textbf{s}^{x'}_{k(2)t},...\textbf{s}^{x'}_{k(N_k)t})$ and $\textbf{S}^{y'}_{kt}=stack(\textbf{s}^{y'}_{k(1)t},\textbf{s}^{y'}_{k(2)t},... \textbf{s}^{y'}_{k(N_k)t})$, respectively, where the stacking operation gathering multiple 1D representation vectors as multi-channels for convolution, $N_k$ is the total number of landmarks of the $k$ th group, $\textbf{s}^{x'}_{k(j)t} (1 \leq j \leq N_k)$ and $\textbf{s}^{y'}_{k(j)t}$  denote the 1D representations produced by Eq. \eqref{transxy} for the $j$ th landmark of the $k$ th group. Here we replace the subscript $n (1\leq n \leq N)$ of $\textbf{s}^{x'}_{nt}$  and $\textbf{s}^{y'}_{nt}$ in Eq. \eqref{transxy} with the subscript $k(j)$ just to mark the switch between the universal index of all landmarks to the relative index within a landmark group. We integrate the component-level structural information into the features of the $k (1\leq k \leq 7)$ th landmark group via two residual 1D convolutional layers, as shown in Eq. \eqref{ps_sffm_1}:
\begin{equation}
	\begin{aligned}
		\label{ps_sffm_1}
		\textbf{P}^{x}_{kt} = LN(Conv1D(\textbf{S}^{x'}_{kt})+\textbf{S}^{x'}_{kt})\\
    	\textbf{P}^{y}_{kt} = LN(Conv1D(\textbf{S}^{y'}_{kt})+\textbf{S}^{y'}_{kt})
    \end{aligned} 
\end{equation}
where LN denotes the layer normalization operation. 

Furthermore, to capture global facial shapes, we adopt an inter-group global geometric encoding mechanism based on the output features from the intra-group geometric encoding module. Let $\textbf{P}^{x}_{t}=stack(\textbf{P}^{x}_{1t},\textbf{P}^{x}_{2t},...\textbf{P}^{x}_{7t})$, and $\textbf{P}^{y}_{t}=stack(\textbf{P}^{y}_{1t},\textbf{P}^{y}_{2t},...\textbf{P}^{y}_{7t})$, where the stacking operation gathers the inputs along the channel dimension for convolution. We introduce two residual 1D CNNs to embed global facial structures as a refinement of the 1D representations for each landmark, as shown in Eq. \eqref{ps_sffm_2}:
\begin{equation}
	\begin{aligned}
		\label{ps_sffm_2}
		\textbf{F}^{x}_{t} = LN(Conv1D(\textbf{P}^{x}_{t})+\textbf{P}^{x}_{t})\\
		\textbf{F}^{y}_{t} = LN(Conv1D(\textbf{P}^{y}_{t})+\textbf{P}^{y}_{t})
	\end{aligned} 
\end{equation}
where the $n$ th channels of $\textbf{F}^{x}_{kt}$ and $\textbf{F}^{y}_{kt}$, denoted $\textbf{f}^{x}_{nt}$ and $\textbf{f}^{y}_{nt}$ respectively, are the features for the $n$ th landmark refined by global structural patterns. From the proposed geometric encoding modules, facial geometric patterns are captured through feature communications among different facial landmarks, and the learned facial structures help to correct the disturbances on facial appearances caused by occlusions or uneven illumination conditions, and further improve the landmark tracking performance for ``in-the-wild" scenarios. It is worth noting that, we take 1D convolutional layers instead of attention layers for token communications in the spatial dimension. The reason lies in that, unlike the temporal tokens which are the features of one certain facial landmark from different time steps, the spatial tokens are features from different landmarks which constitute a structure,  and we find that the 1D convolutional layers show a better capability on modeling the general structural patterns among different landmarks while keeping their clear semantic distinguishments.  See supplementary materials for a empirical study.

\subsection{Optimization}
\label{sec_alg}
Our method is trained with the following loss fucntions:
\begin{small}
	\begin{equation}
		\label{all}
		\begin{split}
			L_o =&  \lambda_h L_h +\lambda_c L_c,  \\
			L_h=&\sum_{t=1}^T\sum_{n=1}^N(||\textbf{h}^x_{nt}-\textbf{h}^{x*}_{nt}||_2^2+||\textbf{h}^y_{nt}-\textbf{h}^{y*}_{nt}||_2^2),\\ 
			L_c =& \sum_{t=1}^T\sum_{n=1}^N((||c^x_{nt} - c^{x*}_{nt}||_2^2)+(||c^y_{nt} - c^{y*}_{nt}||_2^2)) \\
		\end{split}
	\end{equation}
\end{small}
where $\lambda_h$ and $\lambda_c$ are the hyper-parameters determining the weights of each loss in the overall loss function; $L_h$ is the training loss for 1D heatmap regression; $\textbf{h}^{x*}_{nt}$ and $\textbf{h}^{y*}_{nt}$ are the heatmap labels, which are 1D Gaussian distributions around the ground truth landmark positions; $L_c$ is the training loss for confidence regression; $c^{x*}_{nt}$ and $c^{y*}_{nt}$ are the confidence labels. Due to the difficulty of manual annotations for confidences on feature qualities, current benchmark databases for facial landmark tracking do not provide such labels.  As an alternative, we have to infer confidence labels through a flow-based model \cite{RLE} which models landmarks' positional certainty reflecting the texture qualities, and take its inference results as the pseudo labels of confidences. For a better generalization, we just take these pseudo labels as early guidance of our confidence branches, then fine-tune the confidence branches through the supervised signals from heatmap regression in an end-to-end manner. Formally, denote the total training epochs as $E$, when $1\leq e \leq E/2$ we train the tracker with $L_o$; when $E/2 \le e \leq E$ we train the tracker with $L_h$. Adam is used for optimization.

\section{Experiments}
In this Section, we first describe the experimental conditions (\ref{ec}), then give the results and analyses of the ablation studies (\ref{as}) as well as the empirical study on the length of temporal window (\ref{esl}), then compare our method with related works (\ref{cp}). Notably, due to the page limits, the details on hyper-parameter selections, the empirical study on different token communication strategies, the discussion on the computational complexities, as well as more visualization comparisons on the tracking results, are attached in the supplementary materials.

\subsection{Experimental Conditions}
\label{ec}
Two video databases, i.e., the 300VW \cite{shen2015first} database and the TF \cite{FGNET} database, are used to evaluate performance of the proposed method. The 300VW database, i.e., the most widely-used database in the field of facial landmark tracking, contains  $114$ ``in-the-wild" videos, each frame is annotated with $68$ facial landmarks. $50$ videos are divided as the training set and $64$ videos for testing. The testing set can be further categorized as three subsets, i.e., S1, S2  and S3, according to their challenging levels.  

To be consistent with the experimental conditions of previous works \cite{TSTN,AOHR}, for experiments on the 300VW database, both the 300VW training set and the 300W \cite{300w} training set is used for training. The 300W database is an image database with 3,148  training images, without any temporal information. We just make multiple copies of an image to be a pseudo static video to pre-train our tracker, then fine-tuning it on the 300VW training set. We adopt all of the $68$ landmarks when testing on the 300VW database, and adopt $7$ landmarks common to the 300VW and the TF databases for testing on the TF database.  All faces are cropped from the facial bounding boxes and resized to $256\times256$ pixels, then fed into the network for training. Details on hyper-parameter selection are attached in supplementary materials.

We evaluate the tracking performance on both accuracy and stability. Accuracy reflects the closeness of the predicted landmark coordinates to the ground truths. We use \textit{N}ormalized \textit{R}oot \textit{M}ean \textit{S}quared \textit{E}rror (NRMSE) \cite{300w} as the accuracy metric. A lower value of NRMSE corresponds to a better accuracy performance.
Stability reflects the consistency of movement between predicted landmarks and ground truths. The stability error \cite{tai2018towards} is defined as the error of landmark displacement between the tracking results and the ground truths. A lower value of the stability error corresponds to a better stability performance.

\begin{table}[h]
	\centering
	\renewcommand\tabcolsep{0.8pt}
	\begin{tabular}{*{9}{c}}
		\toprule
		\multirow{2}{*}{Settings} &\multicolumn{2}{c}{300VW S1}&\multicolumn{2}{c}{300VW S2}&\multicolumn{2}{c}{300VW S3}
		&\multicolumn{2}{c}{TF} \\ \cline{2-9}
		& N & S &N &S & N & S & N & S \\
		\midrule
		$BL$&  3.31  &  0.90  &   3.45 &  0.93 &   4.42  &  1.85  &    2.02  &  0.66 \\ 
		\midrule
		$BL$+$TE^{-a}$& 3.26   &  0.87 &  3.40   &  0.91 &   4.39  &  1.84 &   2.02  & 0.65  \\ 
		$BL$+$TE^{-r}$& 3.06   & 0.80 & 3.01  & 0.76 &  4.10  &  1.61 & 1.98   & 0.57 \\ 
		$BL$+$TE^{-c}$&  3.10  & 0.80 & 3.03  & 0.77 & 4.14   & 1.65 & 2.00  &  0.61 \\ 
		$BL$+$TE$& 2.91   & 0.74  &  2.93  & 0.73  &  3.98  & 1.52  &  1.92   & 0.47 \\  
		\midrule
		$BL$+$IT$&  3.13  & 0.84  & 3.06  & 0.79 & 4.20   &  1.70 &  2.00 & 0.62 \\ 
		$BL$+$IR$&  3.16   & 0.85  & 3.12  &  0.83 & 4.25   & 1.73  &  2.01  & 0.64 \\ 
		$BL$+$IR$+$IT$&  3.09  & 0.82  & 2.99  &  0.76 &  4.12  & 1.64 &  1.98   & 0.58 \\ 
		\midrule
		$BL$+$TE$+$IR$&  2.82  & 0.72  &  2.85 & 0.70 &  3.93  & 1.47  &  1.90   & 0.41  \\ 
		$BL$+$TE$+$IT$&  2.80  &  0.72 &  2.81 & 0.69 &  3.88  & 1.44  & 1.88  &  0.38 \\ 
		$BL$+$TE$+$IR$+$IT$ & 2.74 & 0.70 & 2.74  & 0.67 & 3.80  & 1.42  &  1.86  & 0.35 \\
		\bottomrule
	\end{tabular}
	\caption{NRMSE (N$/\%$) and stability error (S$/\%$)  of each experimental group.}

	\label{abla}
\end{table}

\subsection{Abalation Study}
\label{as}
We conduct multiple groups of experiments for ablation studies. The NRMSE and the stability errors of these experiments are recorded in Table \ref{abla}. The experimental settings are as the following:  $BL$ denotes the baseline method, which only preserves the backbone 1D representation encoder as well as the 1D heatmap decoders. This makes a simple facial landmark detection method without any temporal modeling and structural modeling modules; $TE$ denotes a complete preservation of the temporal modeling mechanisms of 1DFormer;  $TE^{-a}$, $TE^{-r}$, and $TE^{-c}$ are three ablation terms from $TE$:  $-a$ replaces the attention mechanism with a simple token mixing communication mechanism, i.e., directly adding the feature from each time step with the average feature of the time window; $-r$ removes the recurrently token mixing strategy and simply takes the output features of the backbone encoder from one temporal window as the inputs for attention mechanisms, instead of delivering long-term historical information recurrently as Eq. \eqref{transxy} does; $-c$ removes confidence branches from our attention mechanism in both training and testing phases. $IR$ and $IT$ respectively denote the proposed intra-group and inter-group geometric encoding mechanisms. Here, $BL$+$TE$+$IR$+$IT$ corresponds to the full method of our work. To offset the randomness in the experiment, we conducted 10 independent repeated experiments and reported the average value for each metric in Table \ref{abla}. Notably, as the random seed is fixed in the experiment, the variance of all metrics from  repeated experiments is less than $0.02$. From Table \ref{abla}, we have the following observations and analyses:

\begin{figure}
	\centering
	\includegraphics[scale=0.7]{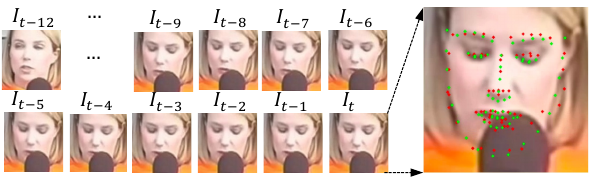}
	\caption{Visualization of the tracking results on a challenging video clip from the 300VW S3. The \textcolor[RGB]{192,0,0}{red points} and \textcolor[RGB]{0,192,0}{green points} are respectively the tracking results without / with the recurrently token mixing strategy.}
	\label{vis_4}
\end{figure}
\begin{figure}
	\centering
	\includegraphics[scale=0.6]{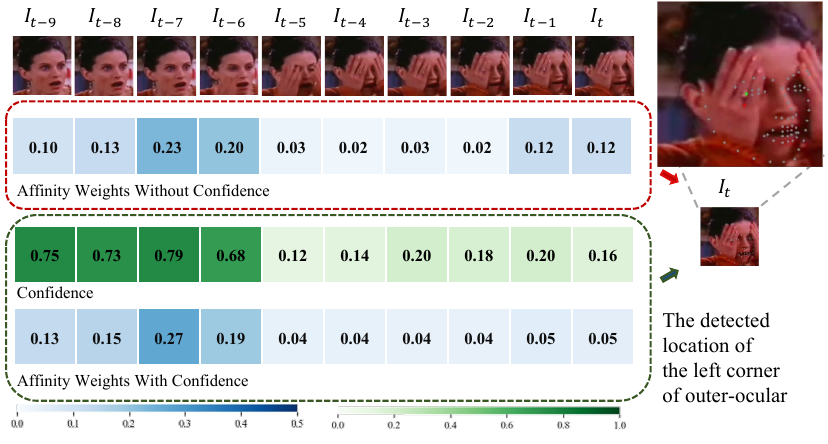}
	\caption{Visualization of the attention weights as well as the tracked results for an occluded landmark, i.e., the left corner of outer-ocular, from a challenging movie clip. The \textcolor[RGB]{192,0,0}{red point} and \textcolor[RGB]{0,192,0}{green point} are respectively the tracking results without / with the help of the confidence branch.}
\label{vis_3}
\end{figure}
\begin{figure}
\centering
\includegraphics[scale=0.53]{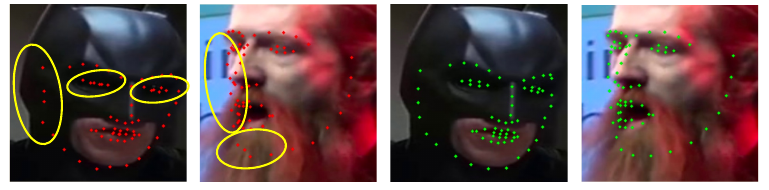}
\caption{Visualization of the tracking results on two challenging video frames, the former from a movie while the latter from the 30VW S3. The \textcolor[RGB]{192,0,0}{red points} and \textcolor[RGB]{0,192,0}{green points} are respectively the results of the tracker without / with structural modeling mechanisms.}
\label{vis_2}
\end{figure}

\begin{figure}[h]
\centering	
\begin{minipage}{0.48\linewidth}
	\centering
	\includegraphics[scale=0.5]{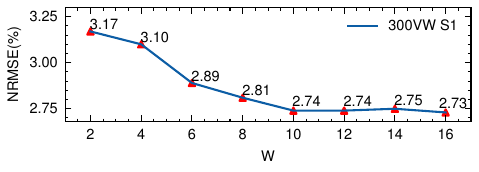}
\end{minipage}%
\begin{minipage}{0.48\linewidth}
	\centering
	\includegraphics[scale=0.5]{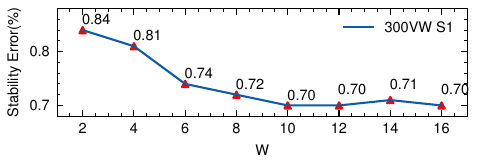}
\end{minipage}%

\begin{minipage}{0.48\linewidth}
	\centering
	\includegraphics[scale=0.5]{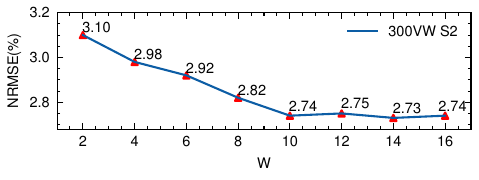}
\end{minipage}
\begin{minipage}{0.48\linewidth}
	\centering
	\includegraphics[scale=0.5]{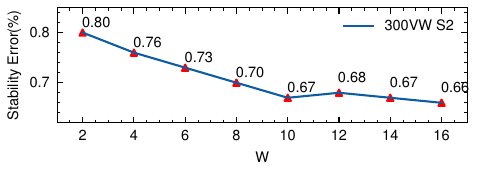}
\end{minipage}

\begin{minipage}{0.48\linewidth}
	\centering
	\includegraphics[scale=0.5]{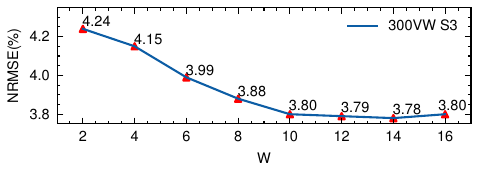}
\end{minipage}
\begin{minipage}{0.48\linewidth}
	\centering
	\includegraphics[scale=0.5]{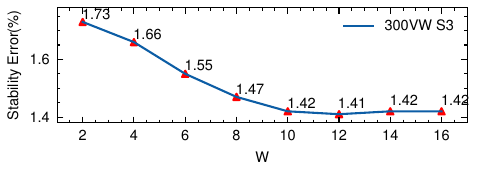}
\end{minipage}

\begin{minipage}{0.48\linewidth}
	\centering
	\includegraphics[width=115 pt, height=45 pt]{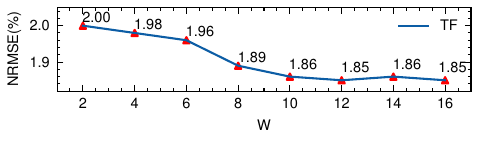}
\end{minipage}
\begin{minipage}{0.48\linewidth}
	\centering
	\includegraphics[scale=0.5]{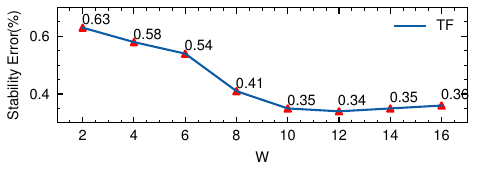}
\end{minipage}
\caption{NRMSE (\%) and stability errors (\%) on 300VW S1, S2, S3, and the TF database with different length ($W$) of the temporal window.}
\vspace{-1mm} 		
\label{fig:w}
\end{figure}

First, comparing between the results of $BL$+ $TE$ and $BL$, we find that the temporal modeling mechanism $TE$ can significantly improve the accuracy and stability performance for facial landmark tracking. This results demonstrate that the proposed temporal modeling mechanisms can effectively integrate the complementary information among multiple video frames to enhance the tracking performance under ``in-the-wild" scenarios. 

Second, the tracking performance of $BL+TE$ shows significant superiority than $BL+TE^{-a}$, demonstrating the effectiveness of the attention mechanism. This is because the attention mechanism can dynamically adjust the weights of feature fusion, enabling it to adapt to a wide range of imaging conditions and facial geometric variations. 

Third, we can observe that recurrently token mixing strategy contributes to the tracking performance by the result that  $BL+TE$ outperforms $BL+TE^{-r}$. In the example of Figure \ref{vis_4}, without the proposed strategy, the tracking performance at the $t$ th frame is very poor since all of the frames in the temporal window, i.e., from $t-9$ to $t$ for a window length of $10$, suffer from severe occlusions; in contrast, if we add this strategy, we can deliver informative features out of the current window. In this case, information delivered from historical frames, e.g., the $t-12$ th frame without heavy occlusions, may help to improve the robustness of the tracking performance. 

Fourth, the contribution of the confidence branch to the attention mechanism is demonstrated by the comparison between $BL+TE$ and $BL+TE^{-c}$ in Table \ref{abla}.  Figure \ref{vis_3} visualizes the attention weights given by $BL+TE$ and $BL+TE^{-c}$  for an occluded facial landmark, namely the left corner of outer-ocular in this case, from a challenging movie clip. We can find that the confidence branch helps the attention mechanism to assign high attention weights if the landmark is visible and assign low weights if it is occluded, providing a plausible guidance to fuse features from different time steps, and thus promoting the localization performance for the occluded frames. 

Fifth, we find that both of the intra-group ($IR$) and inter-group ($IE$) geometric encoding mechanisms promote the tracking performance. Combined with the illustration in Figure \ref{vis_2}, we can find  significant improvement brought by our structure modeling modules. The reason lies that, the $IR$ captures the structure patterns of a local component while the $IE$ models the global face patterns, enhancing robustness of 1D landmark representations to challenging imaging conditions. 

\subsection{Empirical Study on the Length of Temporal Window}
\label{esl}
We make an empirical study on the length of the temporal window ($W$) in the recurrently token mixing strategy. From Figure \ref{fig:w}, we have the following observations. First, when increasing $W$ from 2 to 10, the tracking accuracy and stability boost significantly. The reason lies that, when $W$ is very small, increasing the value of $W$ can provide necessary temporal information for the Transformer to capture. Second, the tracking performance quickly goes to a converge with the increasing of $W$. For example, when $W$ is $10$, the results have no significant gap with $W=16$. The reason may be that, as $W$ reaches an appropriate value, e.g., $10$ in our experiment, the recurrent information delivery strategy from window to window can play an effective role in capturing the long-term temporal patterns of landmark sequences. Thus, there is no need to continuously increase the value of $W$ and a moderate value of $W$ may also bring a good tracking performance comparable to a large $W$. This is an advantage which allows for an acceptable computational complexity of our method.


\subsection{Comparison with Related Works}
\label{cp}
We compare our method with the state-of-the-art methods, which include coordinate regression methods, e.g., TSCN \cite{TSCN},   TSTN \cite{TSTN},  GAN \cite{GAN_Tracking}, and MSKI \cite{MSKI}; 2D heatmap regression methods, e.g., FHR \cite{tai2018towards}, FHR+STA \cite{tai2018towards}, ADC \cite{ADC}, SCPAN \cite{SCPAN}, and SAAT \cite{SAAT}; 1D heatmap regression methods, e.g., the Tracker based on Attentive One-dimensional Heatmap Regression (T-AOHR) \cite{AOHR} and SimCC \cite{SimCC}; as well as the hybrid method of 2D and 1D heatmap regression, i.e., HybridMatch (HM) \cite{HybridMatch}. From these methods, FHR, ADC, SCPAN, SAAT, SimCC, MSKI, and HM are detection methods, which only consider spatial modeling from a static image or video frame;  TSCN, TSTN, FHR+STA, GAN, and T-AOHR are tracking methods which consider both spatial and temporal modeling of a video clip.
\begin{table}[h]
\centering
\renewcommand\tabcolsep{1.2pt}
\begin{tabular}{*{10}{c}}
	\toprule
	\multirow{2}{*}{Method}&\multirow{2}{*}{year}  &\multicolumn{2}{c}{300VW S1}&\multicolumn{2}{c}{300VW S2}&\multicolumn{2}{c}{300VW S3}
	&\multicolumn{2}{c}{TF}\\\cline{3-10}
	& & N & S &N &S & N & S & N & S \\
	\midrule
	TSCN &2014 &	12.54&	-&	7.25&	-&	13.13&	-&	-&	-\\
	TSTN  & 2018 &	5.36&	-&	4.51&	-	&12.84	&-&	2.13&	-\\
	FHR & 2019 &4.82&	2.67&	4.23&	1.77&	7.09&	4.43&	2.07&	0.97\\
	FHR+STA & 2019 & 4.21&	1.58&	4.02&	1.09&	5.64&	2.62&	2.10&	0.69\\
	GAN &	2019 &3.50&	0.89&	3.67&	0.84&	4.43&	1.82&	2.03&	0.59\\
	T-AOHR & 2020 &3.06&	0.84	&3.17&	0.87&	4.12&	1.78	&1.97&	0.64 \\
	SCPAN & 2021 & 4.49 & -& 4.23 & -&5.87 & -& -&- \\
	SAAT& 2021 &3.46&	-&	3.41&	-&	5.23&	-&	-&	-\\
	SimCC  & 2022 & 4.07&  1.22 & 4.12& 1.09  & 5.13 & 2.01  & 2.09 &  0.73 \\
	MSKI  & 2022 & 3.89 & 1.49 & 3.94 & 1.23 & 5.07 &  2.07 & 2.19 & - \\
	HM  & 2023 & 3.15 &  0.89 & 3.23 &  0.92 & 4.28 &  1.80 & 1.99 &  0.67 \\
	\midrule
	Ours & 2024 &\textbf{2.74}   &  \textbf{0.70} & \textbf{2.74}  & \textbf{0.67}  &  \textbf{3.80}  & \textbf{1.42} & \textbf{1.86}  & \textbf{0.35}\\
	\bottomrule
\end{tabular}
\caption{NRSME (N /\%) and stability error (S /\%) of the proposed tracker and the compared methods on the 300VW and the TF databases.}
\label{tab:cp}
\end{table}

Table \ref{tab:cp} lists the NRMSE performance and stability errors of the proposed tracker and the compared methods on the 300VW and the TF database, respectively. The evaluation results of the compared methods are directly copied from literatures, except for SimCC and HM. Due to the lack of published results of these two methods on the respective databases, we just re-implement them under the same experimental condition as our method.  Experimental results from Table \ref{tab:cp} show that our method outperforms the compared methods on both tracking accuracy and stability. Our method outperforms the NRMSE of T-AOHR, which achieves the best overall performance among the compared methods, by 10.46\%, 13.56\%, 7.77\%, and 5.58\% on the 300VW scenarios 1, 2, 3, and TF databases, respectively; we also outperforms T-AOHR at the stability performance by 16.67\%, 22.99\%, 20.22\%, and 45.31\% on the respective databases. Although T-AOHR applied 1D representations on the facial landmark tracking task and achieved remarkable performance, it ignored a deep exploration on temporal and structural modeling for multiple landmarks, as analyzed in Section \ref{related_work}. We address these weaknesses by developing a Transformer architecture to release the good potentials of 1D representations on modeling the long-range sequential  patterns as well as the local and global geometric patterns of facial landmarks, thus achieving significant performance boosts compared to T-AOHR and other related works. We also make visualization comparisons of our method and the related works by rendering their tracking results, please see supplementary materials for details.

\section{Conclusions}
On addressing the weaknesses of current 1D heatmap regression methods and fully exploring the good potentials of 1D landmark representations on temporal and structure modeling of multiple facial landmarks, we propose a facial landmark tracking method based on a new Transformer architecture, namely 1DFormer, to model long-range temporal patterns as well as the local and global facial structures via token communications in the temporal and spatial dimensions, respectively. Specifically,  a confidence-enhanced multi-head attention mechanism with a recurrently token mixing strategy is proposed for temporal modeling; an intra-group and an inter-group geometric encoding mechanism are presented for structure modeling. Experimental results on the 300VW and the TF databases demonstrate that our method achieves state-of-the-art performance for facial landmark tracking with a good modeling capability on the temporal dynamics as well as the geometric patterns of facial landmarks. 

\bibliographystyle{named}
\bibliography{egbib}
\newpage
\appendix
\onecolumn

\section{Feature Fusion Steps of the Confidence-Enhanced Multi-Head Attention Mechanism}
At the end of Section 3.2 of the main body of this paper, the feature fusion steps of the confidence-enhanced multi-head attention mechanism are not described in detail due to the page limits. Here we give a step-by-step description of the feature fusion operations through Eq. \eqref{eq:bs3}:

\begin{equation}
	\label{eq:bs3}
	\centering
	\begin{aligned}
		\textbf{V}_{nt}^{xh} &= \textbf{I}_{nt}^{x}{\textbf{W}_v^{xh}}^T, \,\,\, \textbf{V}_{nt}^{yh} = \textbf{I}_{nt}^y{\textbf{W}_v^{yh}}^T\\
		\textbf{s}_{nt}^{xh'} &= \textbf{a}_{nt}^{xh} \textbf{V}_{nt}^{xh}, \,\,\, 
		\textbf{s}_{nt}^{yh'} = \textbf{a}_{nt}^{yh} \textbf{V}_{nt}^{yh} \\
		\textbf{o}_{nt}^{x} &= \textrm{Concat}(\textbf{s}_{nt}^{x1'}  , \textbf{s}_{nt}^{x2'}  ,..., \textbf{s}_{nt}^{xH'}){\textbf{W}_{o}^x}^T\\
		\textbf{o}_{nt}^{y} &= \textrm{Concat}(\textbf{s}_{nt}^{y1'}, \textbf{s}_{nt}^{y2'}  ,..., \textbf{s}_{nt}^{yH'}){\textbf{W}_{o}^y}^T\\
		\textbf{o}_{nt}^{x'} &= \textrm{LN}(\textbf{o}_{nt}^{x} + \textbf{s}_{nt}^{x}) , \,\,\, \textbf{o}_{nt}^{y'} = LN(\textbf{o}_{nt}^{y} +\textbf{s}_{nt}^{y})\\
		\textbf{s}_{nt}^{x'} &= \textrm{LN}(\textrm{FFN}(\textbf{o}_{nt}^{x'}) + \textbf{o}_{nt}^{x'}) , \,\,\,
		\textbf{s}_{nt}^{y'} = \textrm{LN}(\textrm{FFN}(\textbf{o}_{nt}^{y'}) + \textbf{o}_{nt}^{y'})
	\end{aligned}
\end{equation} 
where $n$ is the landmark index; $t$ is the time step; $h$ is the head index of the attention mechanism; $\textbf{I}^{x}_{nt} = [\textbf{s}^{x}_{nt}; \textbf{s}^{x'}_{n(t-1)};...;$ $\textbf{s}^{x'}_{n(t-W+1)}]$ and $\textbf{I}^{y}_{nt} = [\textbf{s}^{y}_{nt}; \textbf{s}^{y'}_{n(t-1)};...;$ $\textbf{s}^{y'}_{n(t-W+1)}]\in \mathbb{R}^{W \times L}$ denote the input feature sequences for the $t$ th window chunk of the attention mechanism;  $\textbf{a}^{xh}_{nt} \in \mathbb{R}^{1 \times W}$ and $\textbf{a}^{yh}_{nt} \in \mathbb{R}^{1 \times W}$ denote the vectors formed by attention weights;  $\textbf{W}^{xh}_{v} $ and $\textbf{W}^{yh}_{v} \in \mathbb{R}^{d_h \times L}$, as well as $\textbf{W}^{x}_{o} $ and $\textbf{W}^{y}_{o} \in \mathbb{R}^{L \times L}$ are the projection matrices; $\textbf{V}^{xh}_{n}$ and $\textbf{V}^{yh}_{n} \in \mathbb{R}^{W \times d_h}$ are the value matrices; LN is the layer normalization operation; FFN denotes a two-layer feed forward networks.

\begin{figure}
	\centering
	\includegraphics[scale=0.25]{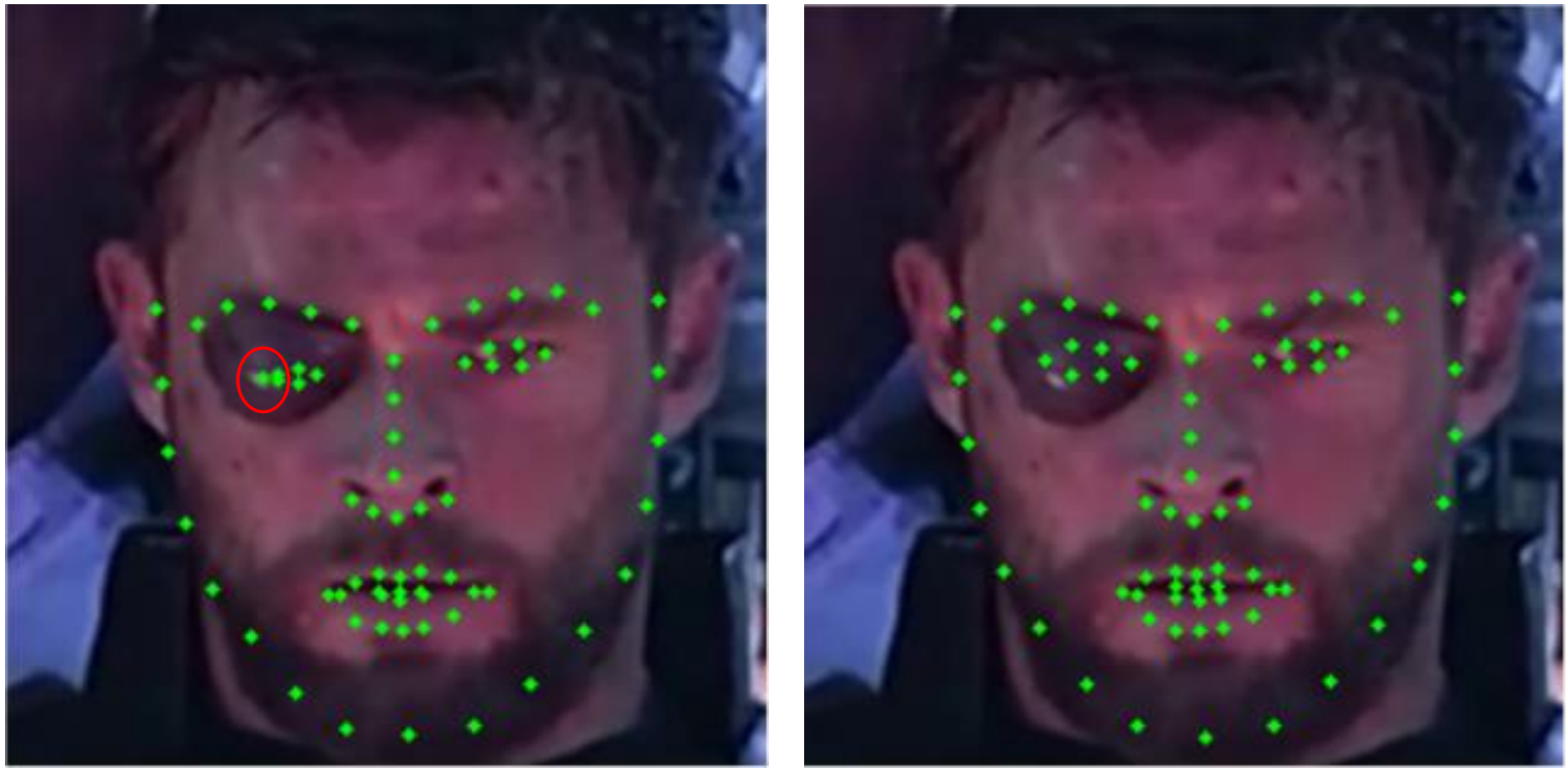}
	\caption{Visualization of the tracking results by different token communication stategies on a challenging movie frame. In the left image, we show the results of $T\_Att\&S\_Att$, while in the right image, we show the resuls of $T\_Att\&S\_Conv$. Please refer to the texts in Part \ref{tcs} for their definitions. In the left image, the tracker is somewhat confused about the semantics of two landmarks in the occluded eye area and the predicted landmark positions are wrongly squeezed to be close to each other; while in the right image, the landmarks in the occluded area can be better inferred with clear semantic distinguishments.}
	\label{vis_0}
\end{figure}

\begin{table*}[!htb]
	\centering
	\renewcommand\tabcolsep{1.8pt}
	\begin{tabular}{*{9}{c}}
		\toprule
		\multirow{2}{*}{Settings} &\multicolumn{2}{c}{300VW S1}&\multicolumn{2}{c}{300VW S2}&\multicolumn{2}{c}{300VW S3}
		&\multicolumn{2}{c}{TF} \\ \cline{2-9}
		& N & S &N &S & N & S & N & S \\
		\midrule
		$T\_Att\&S\_Att$ &  2.87  &  0.74 & 2.86 & 0.72 &  3.91  & 1.49  & 1.90  &  0.44 \\ 
		$T\_Conv\&S\_Conv$ & 2.92  &  0.78 &  2.94 & 0.75 &  4.02  & 1.58  & 1.97  &  0.56 \\
		$T\_Conv\&S\_Att$ &   2.91 & 0.76 & 2.92  & 0.74 & 3.99  & 1.53  &  1.95  & 0.51 \\ 
		$T\_Att\&S\_Conv$ (Ours) & \textbf{2.74} & \textbf{0.70} & \textbf{2.74}  & \textbf{0.67} & \textbf{3.80}  & \textbf{1.42}  &  \textbf{1.86}  & \textbf{0.35} \\
		\bottomrule
	\end{tabular}
	\caption{NRMSE (N$/\%$) and stability error (S$/\%$)  on different token communication strategies.}
	\label{abla_token}
\end{table*}
\begin{table*}[!htb]
	\centering
	\begin{tabular}{*{5}{c}}
		\toprule
		\multirow{2}{*}{Sub-moudles} & Backbone 1D  & 1DFormer & \multirow{2}{*}{Decoder}  & \multirow{2}{*}{All}  \\ 
		& representation encoder \cite{AOHR}&blocks&  &  \\
		\midrule
		Params & 29.73M & 6.88M & 1.58M & 38.19M \\
		GMACs & 28.05 & 0.80 & 0.11 & 28.96 \\
		\bottomrule
	\end{tabular}
	\caption{Numbers of parameters and GMACs for each sub-moudle of our facial landmark tracker. We can find that the proposed 1DFormer is slight and only accounts for a small part of parameters and GMACs of the whole tracker.}
	\label{tab:params_and_flops_cp}
\end{table*}
\begin{figure*}[!htb]
	\centering
	\includegraphics[scale=0.9]{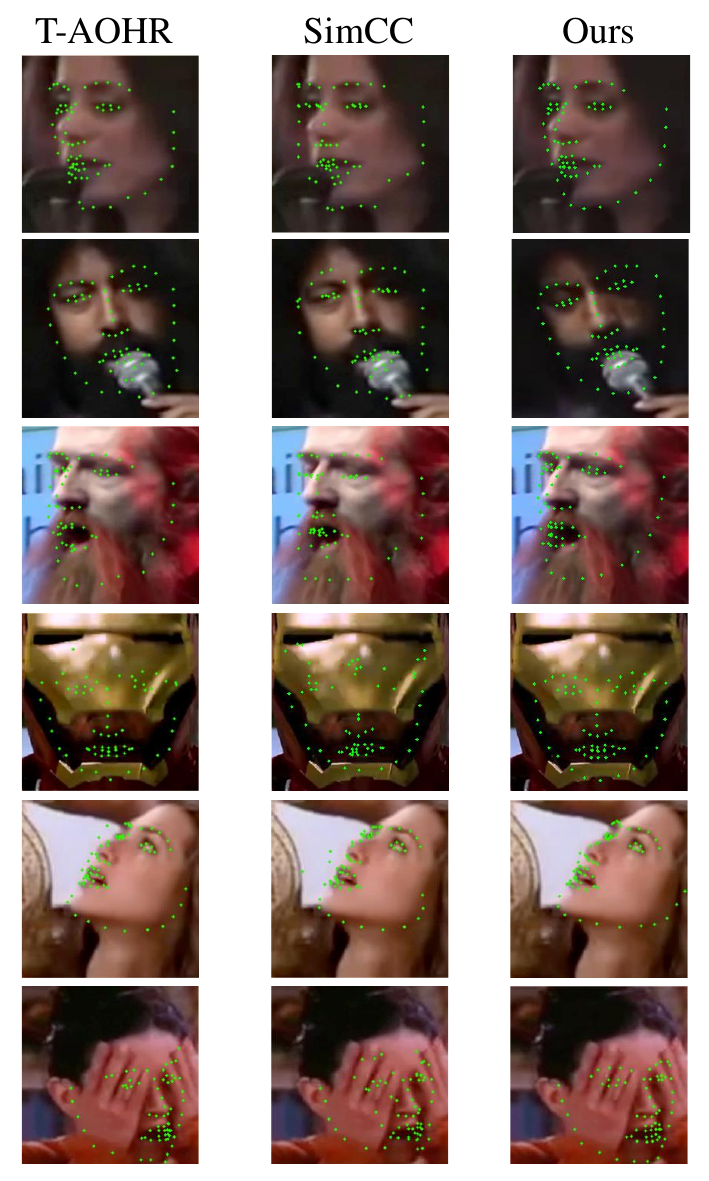}
	\caption{Visualization of the tracking results of different tracking methods, the top is T-AOHR method, the middle is SimCC method, and the bottom is the proposed 1DFormer.}
	\label{vis_1}
\end{figure*}

\section{Details on Hyper-parameter Selection}
The training batch size is set as $10$. Other hyper-parameters, i.e.,  the resolution ($L$) of the 1D features in the hidden stage, the resolution ($D$) of output heatmaps, the input window size ($W$) of Transformers, the block number ($M$) of 1Dformer, the head number ($H$) of attention mechanisms, training epochs ($E$), as well as the training weights ($\lambda_c$ and $\lambda_h$) of loss functions, are determined by cross-validation. We first split the $50$ videos of the 300VW training set as $10$ folds, and conduct $10$-folds cross-validation for parameter search. The search ranges are: $L,D \in \{128, 256, 512, 768, 1024\}$; $W \in \{2, 4, 6, 8, 10, 12,14,16\}$; $M, H \in \{1, 2, 3, 4, 5\}$; $\lambda_c, \lambda_h \in \{0.1, 0.2, ...,$ $ 0.9\}$; $50\leq E \leq 150$. A coarse-to-fine grid search for hyperparameters is conducted. For $E$, $\lambda_c$ and $\lambda_h$, we assign them the average optimal values from the $10$-folds cross-validation; while for $L$, $D$, $W$, $M$, and $A$, considering both the tracking accuracy and the inference efficiency, we assign them the minimum values that the tracking accuracy shows no statistically significant boost (whether $p\_value<0.05$) by value increments. The searched values for these parameters are: $L^{*}=256$, $D^*=512$, $W^*=10$, $M^*=2$, $H^*=4$, $E^*=64$, $\lambda^*_h=0.9$, $\lambda^*_c=0.1$. After cross-validation, we adopt all of the hyper-parameters as their searched optimal values and re-train the proposed method on the whole training set.

\section{Empirical Study on Different Token Communication Strategies}
\label{tcs}
We take attention layers for temporal token communication to capture the dynamic patterns of each landmark, while take 1D convolutional layers for token communication in the spatial dimension to embed the structural patterns of multiple landmarks into their 1D representations. Here we give the experimental results for different token communication strategies on the temporal and spatial dimensions in Table \ref{abla_token} and make analyses. In Table \ref{abla_token},  $T\_Att\&S\_Att$ means taking the attention layers for token communications in both the temporal and the spatial dimensions; $T\_Conv\&S\_Conv$ means taking the 1D convolutional layers for token communications in both the temporal and the spatial dimensions; $T\_Conv\&S\_Att$ means taking the 1D convolution layers for temporal token communication and taking the attention layers for spatial token communication; $T\_Att\&S\_Conv$ means taking the attention layers for temporal token communication and taking the 1D convolution layers for spatial token communication.

From Table \ref{abla_token}, we find that $T\_Att\&S\_Conv$, i.e., the adopted setting in our method, achieves the best results. These results demonstrate that, under the 1D representation learning framework of facial landmarks, the attention layers are more adept at modeling the sequential correlations among temporal tokens, which are the features of one certain facial landmark from different time steps; and 1D convolutional layers show a better capability on capturing the geometric patterns among different landmarks while keeping their clear semantics. As visualized in Figure \ref{vis_0}, compared to the attention layers, when we take 1D convolutional layers for spatial token communications,  the landmarks in the occluded area can be plausibly inferred by the structural patterns of the face and the semantics of different landmarks are clearly distinguished.

\section{Discussion on Computational Complexity}
Table \ref{tab:params_and_flops_cp} shows the numbers of parameters and GMACs\footnote{Evaluated with the \href{https://github.com/Lyken17/pytorch-OpCounter}{thop} tool library.} of each sub-moudle of our tracker. From Table \ref{tab:params_and_flops_cp} we could find that the 1DFormer blocks are very slight compared to the backbone encoder \cite{AOHR} and only accounts for $18.02\%$ and $2.76\%$ of the total parameters and GMACs, respectively. In the main body of this paper, experimental results of abation study demonstrate the effectiveness of 1DFormer, and here the statistics from Table \ref{tab:params_and_flops_cp} also show its efficiency.

\section{Comparisons between 1DFormer and the related works by visualizations}
In the main body of this paper we give the quantitative comparisons of our method and the related works. Here in Figure \ref{vis_1}, we visualize the tracking results of our method and two representative 1D heatmap regression methods, i.e., T-AOHR \cite{AOHR} and SimCC \cite{SimCC}, for a further comparison. From Figure \ref{vis_1}, we can observe that compared to T-AOHR and SimCC, the proposed 1DFormer can perform well under challenging image conditions, such as non-frontal head poses and extreme occlusions that commonly encountered in real-world scenarios.

\end{document}